\documentclass[english]{sobraep}

\usepackage[utf8]{inputenc}
\usepackage{lipsum}
\usepackage{mathtools}
\usepackage{enumitem}
\usepackage{hyperref}
\usepackage{float}
\usepackage{pdflscape}

    \usepackage[T1]{fontenc}
    \usepackage{newtxtext, newtxmath} 
    
    \usepackage{babel} 
    \usepackage[demo]{adjustbox}
   \usepackage{float, caption, makecell, cellspace}
\usepackage{graphicx}
\graphicspath{ {images/} }
\usepackage{algorithm}
\usepackage{threeparttable} 
\usepackage{algpseudocode}
\usepackage{amsmath}
\usepackage{biblatex}

\usepackage{amsfonts}
\usepackage[table,xcdraw]{xcolor}
\usepackage{multicol}
\usepackage{booktabs}
\usepackage{tabularx}
\usepackage{longtable}
\usepackage{vcell}

\DeclareMathOperator*{\argmin}{argmin}

\newcommand{\finalcells}[2]{%
  \begingroup\sbox0{\begin{minipage}{3cm}\raggedright#1\end{minipage}}%
  \sbox2{\begin{minipage}{3cm}\raggedright#2\end{minipage}}%
  \xdef\finalheight{\the\dimexpr\ht0+\dp0+\smallskipamount\relax}%
  \xdef\finalheightB{\the\dimexpr\ht2+\dp2+\smallskipamount\relax}%
  \ifdim\finalheightB>\finalheight
    \global\let\finalheight\finalheightB
  \fi\endgroup
  \begin{minipage}[t][\finalheight][t]{3cm}\raggedright#1\end{minipage}&
  \begin{minipage}[t][\finalheight][t]{3cm}\raggedright#2\end{minipage}}

  \algdef{SE}[VARIABLES]{Variables}{EndVariables}
   {\algorithmicvariables}
   {\algorithmicend\ \algorithmicvariables}
\algnewcommand{\algorithmicvariables}{\textbf{local variables}}

\title{INTEGRATION OF EVOLUTIONARY AUTOMATED MACHINE LEARNING WITH STRUCTURAL SENSITIVITY ANALYSIS FOR COMPOSITE PIPELINES}

\author{Nikolay O. Nikitin$^{1}$, Maiia Pinchuk$^{1}$, Valerii Pokrovskii$^{1}$, Peter Shevchenko$^{1}$, Andrey Getmanov$^{1}$, Yaroslav Aksenkin$^{1}$, Ilia Revin$^{1}$, Andrey Stebenkov$^{1}$, Ekaterina Poslavskaya$^{2}$, Anna V. Kalyuzhnaya$^{1}$ \\
	\normalsize $^{1}$ITMO University, Saint Petersburg, Russia\\
 	\normalsize $^{2}$Huawei Technologies Co., Ltd\\
	\normalsize e-mail: nnikitin@itmo.ru, maiiapinchuk@gmail.com
}
\begin{document}

\maketitle

\section{Abstract}

Automated machine learning (AutoML) systems propose an end-to-end solution to a given machine learning problem, creating either fixed or flexible pipelines. Fixed pipelines are task independent constructs: their general composition remains the same, regardless of the data. In contrast, the structure of flexible pipelines varies depending on the input, making them finely tailored to individual tasks. However, flexible pipelines can be structurally overcomplicated and have poor explainability. We propose the EVOSA approach that compensates for the negative points of flexible pipelines by incorporating a sensitivity analysis which increases the robustness and interpretability of the flexible solutions. EVOSA quantitatively estimates positive and negative impact of an edge or a node on a pipeline graph, and feeds this information to the evolutionary AutoML optimizer. The correctness and efficiency of EVOSA was validated in tabular, multimodal and computer vision tasks, suggesting generalizability of the proposed approach across domains.

\section{Introduction}

AutoML is a promising direction of research for democratization of data science across fields. However, greater predictive performance is often associated with a decreased stability and interpretability of the pipelines \cite{saeed2023explainable}. Explainability and transparency of a solution are both critical for AutoML democratization and its wider use in real world tasks. Therefore, AutoML should prioritize not only the predictive power, but also the simplicity of a pipeline during its construction. Structural sensitivity analysis (SA) is a promising instrument for such pipeline optimization. 

In prior research, SA was used to assess how the uncertainty in the input data or in parts of a system affected the system as a whole \cite{loucks2017system}. We consider an AutoML pipeline as a graph, therefore, the scope of SA in our study is to quantitatively estimate the positive or the negative effect of an edge or a node in a pipeline graph. Throughout the paper, we will consider pipelines as directed computational graphs, consisting of preprocessing components and of machine learning models arranged as a sequence of consecutively executed stages, where the output of one component becomes input for the next. 

This paper discusses the approaches for constructing pipelines with genetic algorithms extended with the SA module, and compares the quality of the proposed solution with other AutoML systems in terms of their robustness and interpretability. 

The module comprises two algorithms, performing the local and the global SA. The former seeks to simplify the pipeline structure given a fixed metric value. Thus, the local SA evaluates the contribution of each preprocessor or model to the final metric, suggesting redundant pipeline components. The global SA algorithm selects a more efficient pipeline structure based on accumulated meta-knowledge. For example, it proposes operations that have worked best in prior tasks.

We structured the paper as follows: \hyperref[sec_rl]{Section III} gives an overview of the existing algorithms, \hyperref[sec_sa]{Section V} provides the description of the proposed EVOSA module, \hyperref[sec_exp]{Section VI} covers the experimental setup and the benchmarking results, \hyperref[sec_disc]{Section VII} summarizes the experimental findings and discusses future improvements, \hyperref[sec_concl]{Section VIII} highlights the key fundamental and empirical points of the paper.

\section{Related work}
\label{sec_rl}
\label{subsec_target1}
\label{subsec_target6}

This section analyses the historical and the state-of-the-art AutoML solutions, focusing on the underlying algorithms of the pipeline construction. We discuss only open-source frameworks and libraries, since the real effectiveness and internal structure of proprietary AutoML solutions can not be assessed directly. 

\subsection{Existing AutoML approaches}

In this paper, we group the existing AutoML systems based on their pipeline construction approach into solutions that create fixed (linear) pipelines or flexible (variable-shaped) pipelines \cite{nikitin2022automated}. The former result in a standard sequence of stages (preprocessors followed by estimators) invariable across tasks. For example, we preprocess the data with a standard scaler for continuous variables, and with one-hot encoding for categorical variables; then we fit a range of models to the transformed dataset, and either select the most accurate model, or orchestrate all trained models in an ensemble (via stacking or simple averaging). In contrast, variable-shaped pipelines have high structural variability depending on the dataset: they create linear or tree-like structures individually tailored to a given task. For example, the same dataset is preprocessed in several ways, applying feature selection, feature generation, a variety of scaling or encoding schemes, these versions are fed to the estimators, which may follow one another (the output of the first estimator serves as an input for the second), often resulting in tree-like structures with multi-level interconnections (see {\ref{fig_localsa}}).

Some of the earliest attempts to develop AutoML with linear pipelines were Auto-Weka \cite{kotthoff2019auto,Thornton2013} and auto-sklearn \cite{feurer2015efficient}, sharing the same Bayesian optimizer to discover a perfect combination of feature preprocessors, ML models and their  hyperparameters to maximize the predictive performance on unseen data. The fixed pipeline structure from auto-sklearn allows several data preprocessing operators, one feature prepossessing operator, and a single model. Additionally, in auto-sklearn 2.0 simple pipelines can be ensembled together \cite{Feurer2020}. AutoWeka uses a similar approach to pipeline construction and model ensembling, but with a wider range of  base models and hyperparameter configurations. 

A more recent solution, creating a linear pipeline, and demonstrating state-of-the-art performance is H2O AutoML\cite{Ledell2020}. Some H2O AutoML competitiveness can probably be attributed to the high quality of their base models. Unlike  many other AutoML solutions, which rely on the existing open source libraries, H2O developed high-performing versions of the well-known algorithms. For example, XGBoost, Random Forest, Extremely Randomized Trees, and so on. It creates a fixed pipeline for handling any dataset: the base estimators are fitted to the data after they perform all the required preprocessing; these fitted estimators can be ensembled together via stacking. The stacking algorithm includes either all fitted base models, or a single copy of each model type.  Despite its simplicity, the resulting linear pipeline demonstrates a competitive benchmark performance.

The first  AutoML solution designing variable-shaped pipelines was TPOT \cite{olson2016tpot}, which used multi-objective genetic programming, treating the parameter configuration optimization as a search problem. ML operators in TPOT are defined as genetic programming primitives. These primitives are combined into a tree-based pipeline. Models come from the scikit-learn library \cite{pedregosa2011scikit} and belong to one of three groups: supervised classification operators, feature preprocessing operators, and feature selection operators. The optimization starts with the generation of 100 random pipelines. Only the top 20 (based on the cross-validation metric score and the pipeline size) are selected for the next generation. During the optimization, TPOT creates multiple copies of a dataset: each copy can be modified by the operators, these modified outputs can be combined, can be followed by other operators, until they finally become an input for the prediction primitive.

AutoGluon \cite{erickson2020autogluon} is a more recent  AutoML solution, creating variable-shaped pipelines and demonstrating state-of-the-art  performance. Their system supports model orchestration in multi-layered stacking ensembles with multi-layer skip-connections, which connect the output of one layer and the input of another non-adjacent layer. The framework can automatically preprocess  raw data, identify the problem type (binary, multiclass classification, or regression), partition data into folds for model training or validation, fit base models, and create an optimized ensemble model that typically outperforms any fitted base model. The algorithm fits base models in the following predefined order: AutoGluon-Tabular Deep neural networks, LightGBM, CatBoost, Random Forest, Extremely Randomized Trees, and K-nearest Neighbors. The AutoGluon-Tabular AutoML was the first system introducing a multi-layer stack ensemble with skip-connections: better generalization performance and reduced overfitting are achieved by the internal k-fold cross-validation.

LightAutoML, or LAMA \cite{vakhrushev2021lightautoml}, extends the multi-layered pipeline approach implemented in AutoGluon by adding presets, which the authors define as fixed strategies for building dynamic pipelines. These are sequences of stages that have demonstrated their effectiveness in prior tasks.  LAMA enables a lot of customization options: for example, the user can predifine pipelines that will be incorporated in the AutoML assembler. 

One of the most recent frameworks, FEDOT  \cite{nikitin2022automated}, prioritizes maximal variability in candidate pipelines compared to alternative solutions. Similarly to TPOT, FEDOT AutoML uses evolutionary optimization; however, its pipelines are considered as directed acyclic graph-based structures, where nodes represent data transformers and models, and edges indicate data flows. Additionally, FEDOT can handle many data types (including, time series, text, and image data) and problems (both classification, regression). FEDOT uses Pareto optimization (using predictive quality and number of models as concurring objectives) to limit the complexity of the resulting pipelines, which tend to grow overly complex under the traditional evolutionary approach. However, this does not completely resolve the problems of complicated pipelines, as well as long convergence times for some datasets. The structure of a pipeline could be simplified by incorporating a system component that gives quantitative  cost-benefit estimation for each node in a pipeline graph, such as the local SA operator. The convergence time could be significantly decrease by using meta-knowledge from prior tasks, regarding which operations tend to work best, such as via the global SA component. 

To summarize, we discussed two main groups of approaches to AutoML system design. The first group tends to create fixed, linear pipelines that provide a solid, general solution to any given task. The second group seeks individually tailored solutions for given problems, resulting in dynamic, variable-shaped pipelines. The downside of the second approach is a possible excessive complexity and long convergence of AutoML systems. This is particularly relevant for frameworks with evolutionary optimizers, such as TPOT and FEDOT. The rest of the paper focuses on the effects of adding EVOSA to the novel evolutionary AutoML framework FEDOT, where we expect to observe the greatest gains from using both local and global SA. We predict that local SA will increase the explainability and transparancy of the resulting pipelines, while global SA will enhance the convergence times via meta-learning.  

\subsection{Explainability of AutoML}

The term 'explainability' covers different aspects of an AutoML solution: It includes understanding the reasoning underlying the decisions of ML models \cite{du2019techniques}, as well as the choice of a specific pipeline by an AutoML framework \cite{xanthopoulos2020putting}. For the purposes of this study, we will mostly focus on the pipeline transparancy and explainability, because possible obscurity of model decision criteria are not reserved to AutoML. 

The transparency and explainability of pipeline design (so-called "white-box AutoML" \cite{das2020amazon}) are  important problems for integration of AutoML in production workflows.  The researches have proposed different approaches to justification and clarification of the  final pipeline selection \cite{drozdal2020trust}. For example, the recently proposed PipelineProfiler provides an interactive analysis of pipeline search across AutoML systems: Its matrices - where rows represent the various pipelines listed from best to worst, and the columns indicate if the respective pipeline component was used - allow identifying patterns in the data that have been shown useful even for experts \cite{ono2020pipelineprofiler}. Alternative methods of dimensionality reduction are particularly helpful for evolutionary approaches to pipeline design \cite{de2019analysis}.

In some papers, explainability and interpretability of AutoML are associated with specific features useful for a data analyst \cite{xanthopoulos2020putting}, such as data visualization, progress report, final model interpretation, feature importance, and analysis exploration. The underlying idea is that "AutoML should automate, not obfuscate" \cite{xanthopoulos2020putting}, that is, an analyst can check the appropriateness of the proposed AutoML solution, while at the same time get a higher-level perspective on the data.

However, the existing AutoML solutions have limited tools for visualization and interpretation of the pipeline design process. It is particularly important for the frameworks which create dynamic variable-shaped pipelines (especially evolution-based, such as TPOT and FEDOT) that prioritize pipeline customization for each task. 

\subsection{Meta-learning in AutoML}

The other way to make AutoML more effective and understandable is meta-learning, which consists in transferring knowledge from prior tasks to the current problem during pipeline optimization.

The AutoML researchers have primarily attempted to achieve this by restricting the search space (for pipelines, models, or hyperparameters) or by prioritizing more successful setups during optimization. The restriction and modification of a search space increases the convergence of an AutoML solution. The search space reduction in AutoML is often based on human-guided machine learning \cite{d2019modeling}; however, the search space can also be refined using knowledge from prior tasks \cite{cambronero2020ams}, and some patterns can be extracted in an automated way  \cite{zoller2021incremental}. For example, useful knowledge can come from previous optimization runs: A pairwise correlation matrix for pipelines' performance in existing experiments identifies effective pipelines for unseen datasets \cite{fusi2018probabilistic}.

A useful offline store of meta-knowledge can be significantly enhanced by the accumulated public efforts. H2O Driverless AI\footnote{\url{https://h2o.ai/platform/h2o-driverless-ai-recipes}} uses the experience gained by the community to solve common ML problems. The AutoML system implements the efficient data handling strategies and assigns them higher priority. This approach may assist the design of initial assumptions for optimization; however, it requires manual curation of the knowledge database and does not guarantee the efficient transfer of best practices from one problem to another.

Auto-sklearn adopts a similar strategy, with an offline training phase on a large pool of tasks. It fits a separate model that identifies datasets with similar characteristics as the target dataset, and proposes successful pipelines and hyperparameters to warm-start the Bayesian optimization \cite{park_hauschild_heider_2021}.

\subsection{Sensitivity analysis for modelling pipelines}
\label{subsec_target2}

In general, sensitivity analysis investigates the influence of uncertainty in the input data or in parts of the system on the behaviour of a system as a whole \cite{loucks2017system}. The various SA techniques are widely accepted in ML and AutoML \cite{saltelli2021sensitivity}. For example, SA algorithms are used to visualize and explore AutoML pipelines \cite{Ono2020} or decrease the optimization time by dropping unpromising pipelines \cite{Gijsbers2018}. In their pipeline optimization capacity, the SA methods look similar to meta-learning. Hence, it is useful to differentiate between the global SA tools, which indeed constitute meta-learning methods, based on prior (offline) experience, and the local SA tools, which optimize pipeline based on evidence from an ongoing training session. The rest of the section will discuss the existing local SA functionalities available in modern AutoML. 

%There are a large number of approaches to performing a sensitivity analysis for input variables. For example, global sensitivity indices method \cite{sobol2001global} can be used when it is feasible to estimate variance. These indices are used to predict the influence of individual variables or their groups on a model output. SA approach developed in this paper is based on One-factor-at-a-time (OAT) method, since it is interpretable and easy to compute \cite{xu2009using}. OAT method allows to evaluate the effect of each of the variables on the result, changing one input variable at a time, keeping others at their baseline. It allows to test the robustness of the system without excessive computing overhead. The other promising way to improve SA results is the involvement of clustering \cite{roux2021cluster} of predictions to estimate the influence of model inputs and structure to the specific clusters.

%However, when applied to the structural analysis of pipeline, most of the approaches require too expensive computation, since the search space is larger. 

AutoGluon proposes an advanced ensemble model selection algorithm based on the local SA information to maximize the final metric. Specifically, they use model selection with replacement, sorted ensemble initialization, and bagged ensemble selection. AutoGluon individually fits various models and combines them into one ensemble that outperforms the individual trained models via stacking/bagging \cite{autogluon2020}. This method is simple and transparent; on the downside, it is computationally expensive \cite{ensembleselectionautogluon2004} during training and inference.

TPOT modification, LTPOT aims to reduce the number of pipelines evaluated on the entire dataset by the selection process so that only the most promising pipelines must be evaluated on the full dataset. It is achieved by evaluating pipelines on a small subset of the data: only pipelines exhibiting good performance on the subset will be
evaluated on the whole dataset \cite{Gijsbers2018}.

LightAutoML uses the realization of Tree-structured Parzen estimation by the Optuna framework for GBM models to compare the parameters fine-tuned during the fitting process and expert hyperparameters. A mixed tuning strategy provides models in the pipeline, which both are blended or stacked, or dropped if it does not increase the final model performance.

%Stacked ensembles, and its variety a Super Learner algorithm \cite{van2007super}, have a fixed two-level structure. The first level consists of  fitted base models, which make their predictions on unseen data. These predictions are then used as a meta-dataset to train second-level model that assigns optimal weights to these predictions, and makes its final call. H2O uses a combination of fast random search and stacked ensembles \cite{h202020}. After training the base models, two Stacked Ensemble models are trained using H2O's Stacked Ensemble algorithm.%

The ideas of local and global SA already have a lot of application in AutoML systems. Notably, incorporating both offline information (via meta-learning) and online data (coming from the current training session) consistently improve system performance  despite the differences in configurations and algorithms, underlying the local and the global SA in specific AutoML frameworks. This paper  explores the effects of both the local (online) and the global (offline) SA knowledge on the pipeline optimization, estimating their contribution to variable-structured pipelines.

\section{Problem statement}
\label{sec_ps}

AutoML is a computationally expensive optimization task, which is highly affected by "the curse of dimensionality". The computational complexity becomes particularly apparent with systems, creating variable-shaped pipelines. The formulation of the optimization task for the design of custom pipelines has the following symbolic representation:

\begin{equation} 
\begin{gathered}
\label{eq_opt}
\boldsymbol{N_{opt}},\boldsymbol{E_{opt}} = \argmin_{(N, E) \in \boldsymbol{S}}{F(P(\boldsymbol{N},\boldsymbol{E}),\boldsymbol{X},\boldsymbol{Y}, T)}, \\
\boldsymbol{S} = R(\boldsymbol{S_{full}}),
\end{gathered} 
\end{equation}
where $P$ is a pipeline design function that accepts nodes $N$ and edges $E$; $S\_full$ is a combinatorial space that includes all correct combinations of the nodes (models and data operations) and edges (data flows between nodes). 

In practice,  $S\_full$ has a very high dimensionality and the computational complexity of the objective function. The $S\_full$ can be reduced to $S$ using meta-heuristic approaches (represented as function $R$). We can search for the optima of $S$ with a variety of numerical optimization methods. The AutoML tasks are usually considered under a specific timeout $T$. The short timeout can cause the identification of a pipeline far from global optima. However,  excessively large timeout can lead to over-complicated pipeline as a result.

The proper balance between the exploration- and exploitation-based behaviours in AutoML pipeline design  remains an open scientific problem \cite{he2021automl}. First, we wish to restrict the search space $S$ as much as possible; second, we seek an effective optimization strategy that avoids expensive evaluations of over-complicated pipelines; third, we wish to retain high predictive power, while reducing the inference time on unseen data.

The SA methods help restrict the search space during optimization \cite{loubiere2016sensitivity} and estimate the importance of different pipeline blocks \cite{barabanova2021sensitivity} or hyperparameters \cite{probst2019tunability} of specific models. 

In this paper, we propose a controllable approach of evolutionary optimization for ML pipelines that reduces the full combinatorial space $\boldsymbol{S_{full}}$ to the reduced space $S$ (described in the Equation~\ref{eq_opt}) using the SA techniques.

\begin{equation} 
\begin{gathered}
\label{eq_opt}
\boldsymbol{S} = R_{sa}(\boldsymbol{N_{opt}}, \boldsymbol{E_{opt}}),
\end{gathered} 
\end{equation}

where $\boldsymbol{S}$ is a restricted search space, $R_{sa}$ is a function that two proposed SA methods: global and local SA. $\boldsymbol{N_{opt}}$ and $\boldsymbol{E_{opt}}$ can be obtained from final or intermediate solutions for pipeline optimization.

The involvement of SA in pipeline design allows performing the following operations: (1) estimating the importance of each block in a pipeline (particularly useful for a data expert); (2) simplifying a pipeline while controlling for its predictive power; (3) improving the AutoML convergence time.

We formulate the following hypothesis for the performance of the SA: 

\textbf{Hypothesis} The local SA allow improving the convergence of optimisation and reduce the structural complexity of solutions without compromising the predictive power of the final solution.

%\textbf{Hypothesis 1} The local SA component  reduces over-complicated pipelines without compromising the predictive power of the final solution.

%\textbf{Hypothesis 2} The global SA component improves the convergence of the evolutionary optimization algorithm without compromising its predictive power.

  %to the modification of pipelines. Since the number of possible changes that can be applied to the current pipeline is large, the most promising solutions in search space $\boldsymbol{S}$ from previous iterations can be assigned with higher probabilities. This scheme can be easily integrated into the mutation state of the evolutionary algorithms. Also, it is possible to generalize it to the meta-AutoML case.

\section{Proposed approach}
\label{sec_sa}

The proposed Evolution with Sensitivity Analysis (EVOSA) approach is implemented as a graph-based evolutionary optimizer for ML pipelines, which comprises four modules, performing the following tasks: (1) evolutionary design of pipelines; (2) local sensitivity analysis of a pipeline structure; (3) global analysis of the evolution history for a set of prior tasks; (4) interactive visualization of SA and evolution results. 

To get the most of the SA approach, we required AutoML with evolutionary optimization. Therefore, the component (1) in EVOSA was procured by the FEDOT system \cite{nikitin2022automated}. Below we describe the components (2)-(4) and their integration into AutoML.

\begin{figure*}
    \centering
    \includegraphics[width=1.0\linewidth]{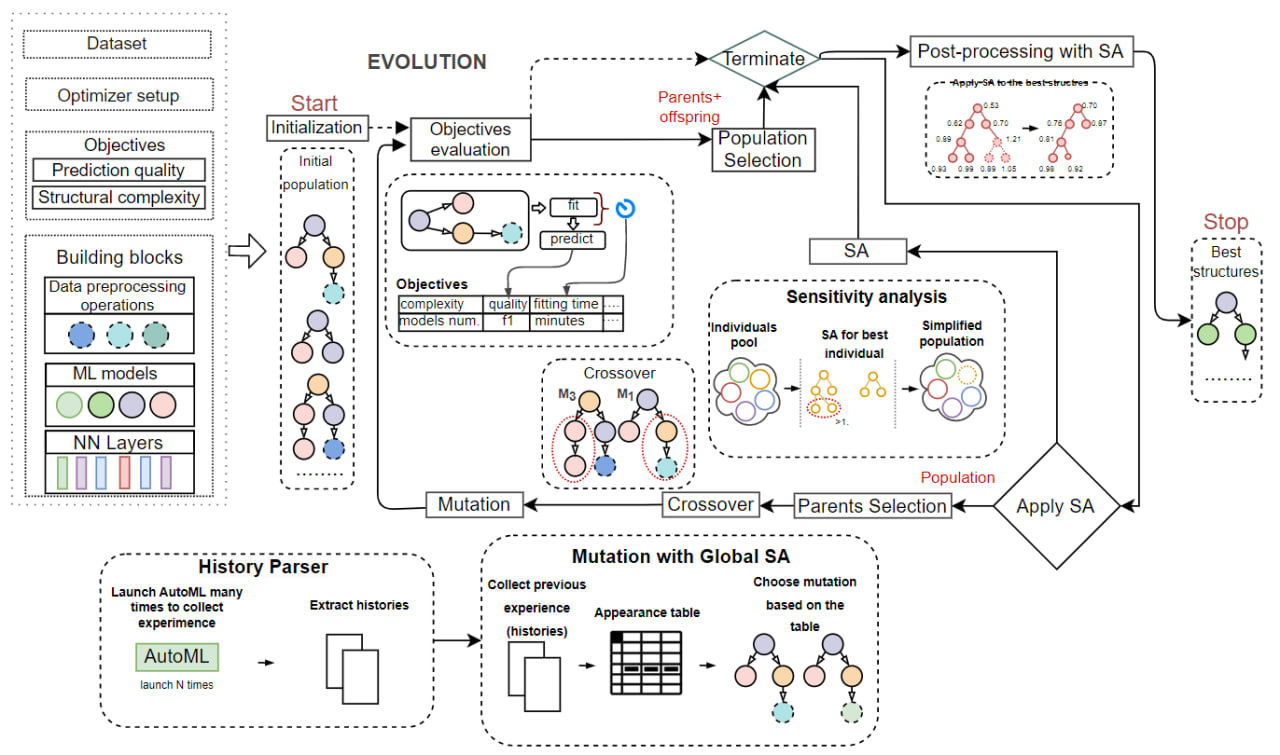}
    \captionsetup{justification=centering}
	\centering
    \caption{The scheme of proposed EVOSA approach.}
    \label{evosa}
\end{figure*}

The schematic overview of the approach is presented in Figure~\ref{evosa}.

\subsection{Local SA component}
\label{subsec_sa_loc}

\begin{figure*}
    \centering
    \includegraphics[width=1.0\linewidth]{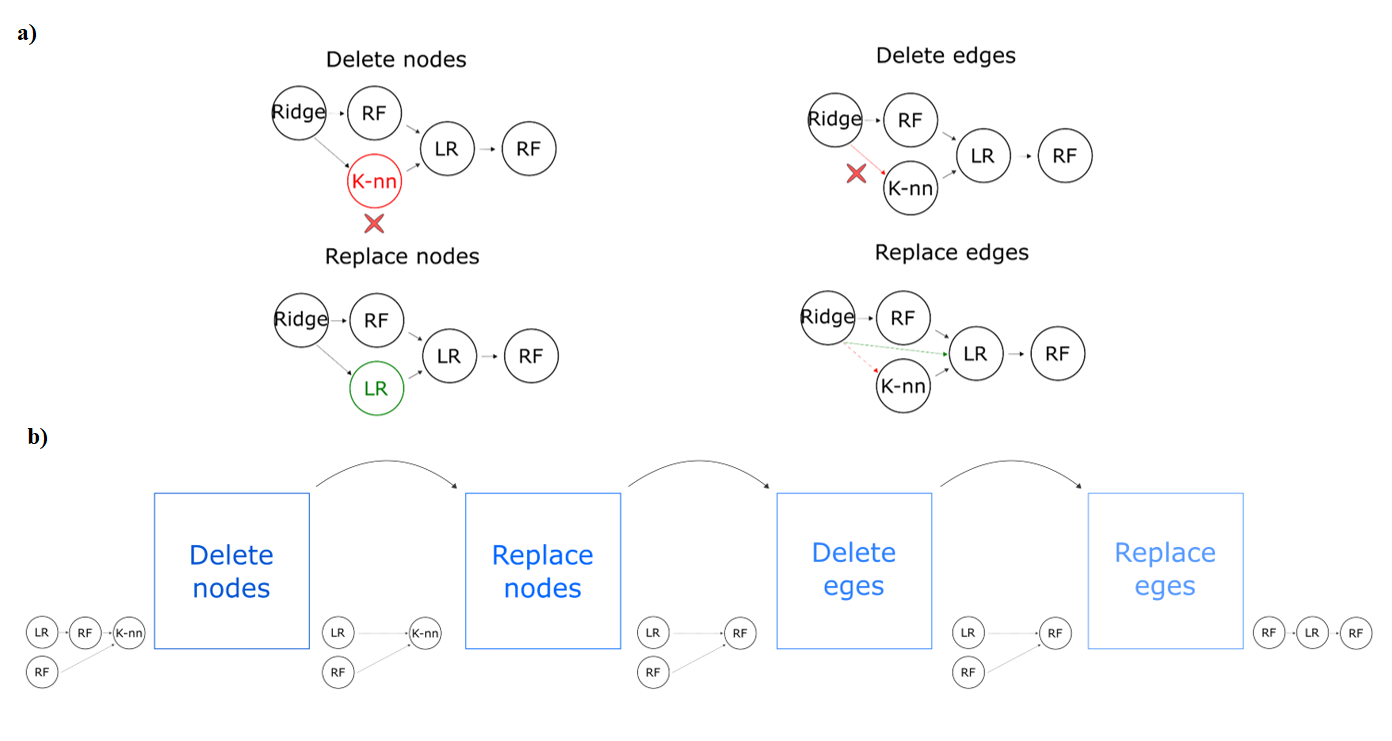}
    \captionsetup{justification=centering}
	\centering
    \caption{Details of the proposed local SA approach: a) example for each variation of operation; (b) the proposed sequence of the pipeline analysis}
    \label{fig_localsa}
\end{figure*}

We consider the pipeline structure as a graph, and the local SA component quantitatively estimates the importance of each node, subgraph and edge.  This information is used to prioritize the replacement and deletion operations during optimization (see Figure \ref{fig_localsa}),

The local SA can assist in multi-objective optimization - by taking into account not only the prediction error, but also the structural complexity of the pipeline, and the training time. For the purposes of this study, the prediction error serves as a primary criterion, the structural complexity and training time - as secondary criteria.

\begin{figure*}
    \centering
    \includegraphics[width=0.55\linewidth]{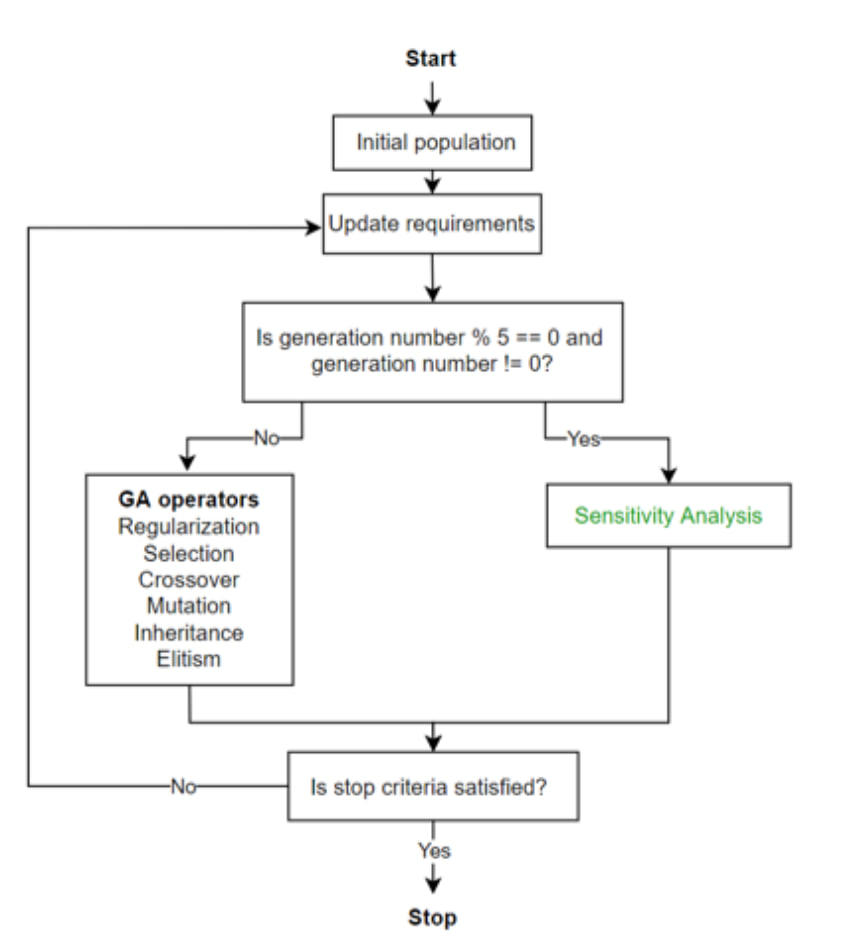}
    \captionsetup{justification=centering}
	\centering
    \caption{Integration of the local SA approach with evolution: SA is applied on $N$ best individuals every $K$-th generation, which allows to improve convergence and get simpler results without metric loss}
    \label{fig_localsa_sch}
\end{figure*}

The SA component functions as mutation operators in the evolutionary optimizer (see the integration scheme in Figure~\ref{fig_localsa_sch}).

% The SA component functions as a post-processing step (optimizers with prefix \textit{PostProc}) and mutation operators (optimizers with prefix \textit{Integrated}) in the evolutionary optimizer (see the integration scheme in Figure~\ref{fig_localsa_sch})

In implemented SA algorithm all graph optimization operations (\textit{Node Deletion}, \textit{Node Replacement}, \textit{Edge Deletion} and \textit{Edge Replacement}) are applied one after another. For example, \textit{Node Deletion} method is applied to all nodes, then the \textit{Node Replacement} method, and so on. 

% We implemented two versions of the SA algorithm: with sequential and with simultaneous execution. In the first version the  graph optimization operations (\textit{Node Deletion}, \textit{Node Replacement}, \textit{Edge Deletion} and \textit{Edge Replacement}) are applied one after another. For example, \textit{Node Deletion} method is applied to all nodes, then the \textit{Node Replacement} method, and so on. In the second version, all methods are applied simultaneously on each iteration, and the combination that gives the most significant metric increase is adopted.

%rewrite
% The local structural analysis can be analyzed using considering the number of the original pipeline ${N}_{total}$ and the number of redundant parts ${N}_{redundant}$ found via analysis. 

% The equation for the pipeline is as follows:

% \begin{equation} 
% \begin{split}
% \label{eq:sustainability_index}
% S_{ind} = \frac{N_{redundant}}{N_{total}}, \\
% \end{split} 
% \end{equation}

Node deletion and node replacement operations allow estimating the node's  redundancy in a pipeline \cite{barabanova2021sensitivity}. Each node's sensitivity index is estimated as follows:

\begin{equation} 
\begin{split}
\label{eq:sensitivity_index_del}
\boldsymbol{S}^{node\_del}_{i} = 1 - \frac{Q(P(N,E))}{Q(P(N\smallsetminus\{N_i\},E'))},  \\
\end{split} 
\end{equation}

where $\boldsymbol{S}^{node\_del}_{i}$ is sensitivity of a vertex during its deletion, $Q$ is the quality measure for the pipeline $P$ formed by a set of nodes $N$ and edges $E$, where $N_i$ is chosen for analysis and thus excluded from the pipeline (value in denominator is quality measure after node deletion). The remaining set of edges after the node removal is denoted as $E'$, since its adjacent edges are also removed.

For the node replacement operation, the following representation is used:

\begin{equation} 
\begin{split}
\label{eq:sensitivity_index_repl}
\boldsymbol{S}^{node\_rep}_{i} = 1 - \frac{Q(P(N,E))}{Q(P(N\smallsetminus\{N_i\}\cup \{\tilde{N_i}\}),E)}, \\
\end{split} 
\end{equation}

where $N_i$ is replaced by new node $\tilde{N_i}$ obtained from search space. %(value in denominator means quality measure after replacement $N_i$ by $N_j$).

For the edge deletion and replacement, the equations are as follows:

\begin{equation} 
\begin{split}
\label{eq:edge_sensitivity_index_del}
\boldsymbol{S}^{edge\_del}_{i} = 1 - \frac{Q(P(N,E))}{Q(P(N, E\smallsetminus\{E_i\}))},  \\
\end{split} 
\end{equation}

\begin{equation} 
\begin{split}
\label{eq:edge_sensitivity_index_repl}
\boldsymbol{S}^{edge\_rep}_{i} = 1 - \frac{Q(P(N,E))}{Q(P(N, E\smallsetminus\{E_i\}\cup E_j)}
%E_i')|E\smallsetminus\{E_i\}\cup E_j)}, \\
\end{split} 
\end{equation}

\subsection{Global SA component}
\label{sec_sa_glo}
The global SA component enhances the performance and stability of the evolutionary algorithms, using accumulated knowledge from previous tasks or runs. In the paper, global SA refers to a specific version of meta-optimization. 

\subsubsection{Appearance analysis}
The structure of pipelines is a directed acyclic graph that can be divided into simple sup-parts - a pair of nodes and an edge between them. The algorithm uses prior history to estimate the importance of all edges in pipelines. The history here is a collection of previous launches that contain information about the pipeline structure and their metric on test samples - fitness score. Based on this information, we can apply heuristics to decide on the suitability of operations. 

These heuristics increase the mutation efficiency by advising the most suitable operations for a graph. The function is calculated for all pairs of operations for that particular score. 
The score between operation $A$ and operation $B$ ($A\rightarrow B$) is defined as 

\begin{equation} 
F_{suit}=\overline{Rn_{A\rightarrow B}},
\end{equation} 

where $\overline{Rn_{A\rightarrow B}}$ is a  mean normalized rank of pipelines with edge  $A \rightarrow B$.

The normalized rank $Rn$ is the position of a pipeline compared to other pipelines based on the fitness divided by the number of pipelines in a history.

\begin{figure}
    \centering
    \includegraphics[width=0.8\linewidth]{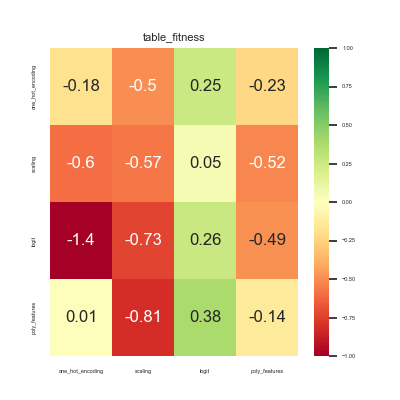}
    \captionsetup{justification=centering}
	\centering
    \caption{Example of a table with suitability score}
    \label{fig_score}
\end{figure}

All values are registered in a table (see visualization in Figure~\ref{fig_score}). Value in a cell can be interpreted as follows: how many times we get an increase in performance if we set an operation from column \textit{X} after the operation from row \textit{Y}.

The AutoML optimizer assigns weights for the operations based on their 'suitability scores' during mutation, thus prioritizing more promising combinations. 
The algorithm is applied to two types of mutations. The first type changes operation in the node. The second type inserts a new node in the graph (as a child, a parent, or an intermediate node). The proposed approach changes only the algorithm for selecting an operation for a new or existing node.

For the node change mutation, the algorithm randomly selects a node to mutate. Next, it finds child and parent nodes for the target node. Finally, it selects the best node based on the information about the child and the parent nodes and the 'suitability values' table.

In case of the addition mutation, the algorithm first randomly selects a place for a node insertion. Second, it identifies new child and parent nodes. Third, it selects the best node based on the information about the child and the parent nodes, and the 'suitability values' table.

The algorithm for selecting suitable nodes selects candidates for mutation and calculates their scores based on the 'suitability values' table. Each candidate receives its score (weight).

The weight for the $i$-th candidate is calculated as a sum of the suitability scores of the child and the parent nodes:

\begin{equation} 
W_i=\sum_{\text{i=1}}^{|\textbf{N{child}}|}F{ij}+\sum_{\text{j=1}}^{\textbf{|N{parent}|}}F{ij},
\end{equation} 

where $\textbf{N{parent}}$ is a set of parent nodes, and the $\textbf{N{child}}$ is a set of child nodes, and $F{ij}$ is the suitability score for $i$-th candidate according to $j$-th parent (child) node this candidate would get if it will be selected.

Candidates with scores below zero are dropped. The final node is selected randomly according to the normalized weights. The pseudocode of the algorithm is described in Alg.~\ref{alg_appearance}.

\subsubsection{Meta-model}
We implemented additional global SA algorithms to improve the results on future runs. We denote $G_i = \{(V_i, E_i)\}$ as a graph, $V_i$ its vertexes and $E_i$ its edges; we denote $F_{(i, d)} = F_d(G_i)$ as a fitness score for the graph $G_i$ on a dataset $d$ and $Rn_{(d, i)}$ as a normalized rank  of the  graph $G_i$ on the dataset $d$.  We represent each graph in the table format. In the proposed approach, we used encoding of all edges as a graph representation: if a graph has two edges "$(A, B)$", then in the column  "$(A, B)$" the row representation will have the value  "2". Let us denote $X$ as a table with representations for each graph and $Y$ as a table with  $Rn_{(d, i)}$ for each graph and dataset. The Cartesian product of $X$ and $Y$ will be our train dataset. Ranks are the target values, and graph representations are the features. We fit a model (in our case, a Random Forest) to predict $Rn_{(d, i)}$ based on the graph representation. 

\begin{figure*}
    \centering
    \includegraphics[width=0.8\linewidth]{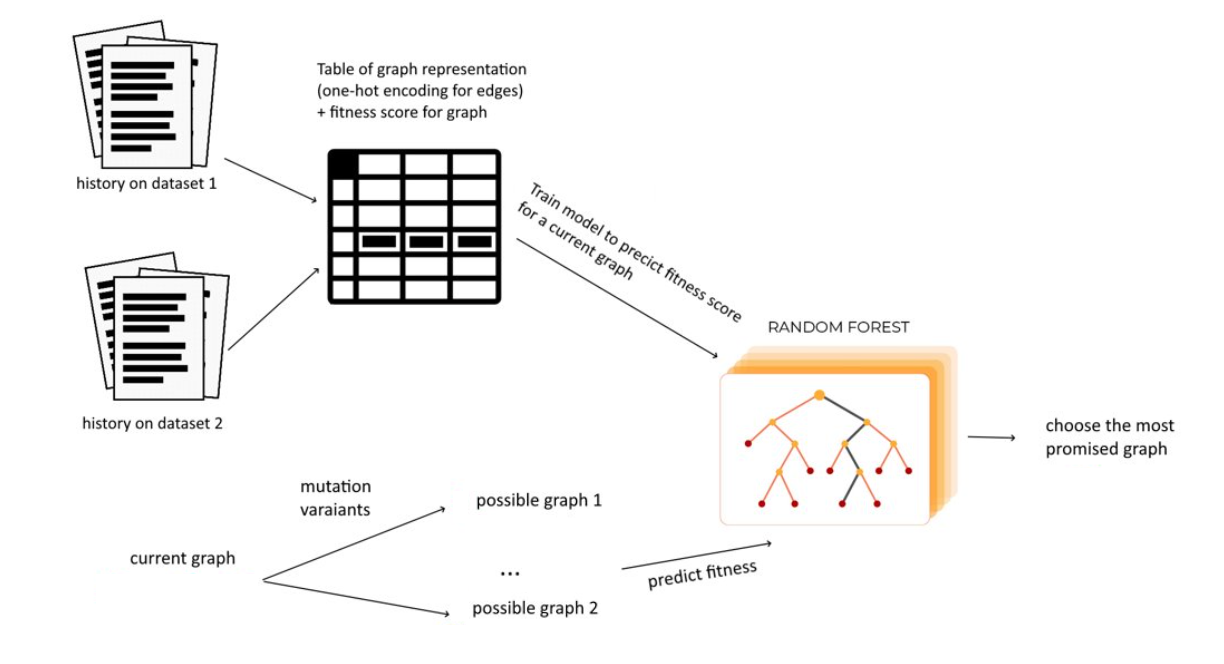}
    \captionsetup{justification=centering}
	\centering
    \caption{Workflow of the proposed approach to global SA (metamodel)}
    \label{fig_metamodel}
\end{figure*}

We use the fitted model for optimization. On each mutation iteration, we get several possible graphs (depending on which node we insert or replace). Now we can calculate $Rn_{i}$ for each graph. This information can help us choose more promising graphs. We randomly select the final graph from the best $N$ graphs based on the $Rn_{i}$. The pseudocode of the algorithm is described in Alg.~\ref{alg_metamodel}. The schema of the approach is presented in Figure~\ref{fig_metamodel}.

\begin{algorithm}[ht!]
\caption{Pseudocode of procedure for selecting node according to suitability table}\begin{algorithmic}[1]
\algnewcommand\And{\textbf{and}}
\algnewcommand\Not{\textbf{not}}

\Procedure{ChooseNodeDirected} {}
    \State \underline{Input:} $parent\_nodes, child\_nodes, candidates,$ \newline$ suitable\_table$
    \State \underline{Output:} $name\_of\_operation$
    %\State $decompose\_operation$ \gets \Call{Repository}{params.task, 'decompose'}
   \State $cand \gets \Call{SET}{candidates, suitable\_table.columns}$
   \State $child \gets \Call{SET}{child\_nodes,\newline suitable\_table.columns}$
   \State $parent \gets \Call{SET}{parent\_nodes,\newline suitable\_table.columns}$
   \State $scores\_parent \gets suitable\_table[parent, cand]$
   \State $scores\_child \gets suitable\_table[cand, child]$
   \State $scores \gets \Call{SUM}{scores\_parent} +\newline  \Call{SUM}{scores\_child}$
   \State $scores \gets \Call{DROP\_BAD\_CONNECTIONS}{scores}$
    \If{$MAX(scores) <$  0.1} 
        \State \textbf{return} $\Call{RAND}{cand}$ 
    \Else
        \State $weights \gets scores / \Call{SUM}{scores}$
        \State \textbf{return} $\Call{RAND}{cand, weights=weights}$  
    \EndIf 
    
\EndProcedure

\end{algorithmic}
\label{alg_appearance}
\end{algorithm}

\begin{algorithm}[ht!]
\caption{Pseudocode of procedure for metamodel mutation (general schema)}\begin{algorithmic}[1]
\algnewcommand\And{\textbf{and}}
\algnewcommand\Not{\textbf{not}}

\Procedure{MutationMetamodel} {}
    \State \underline{Input:} $initial\_graph, candidates, metamodel$
    \State \underline{Output:} $graph$
    \Variables
        \State $op\_scores$, empty dictionary with candidates values as keys 
    \EndVariables
    \For{candidate in candidates}
    \State $template\_graph \gets \Call{MUTATE}{initial\_graph}$
        \State $op\_scores[candidate] \gets \Call{COMPUTE\_SCORE}{\newline template\_graph}$
    \EndFor
   \State $final\_choose \gets \Call{GET\_RANDOM\_FROM\_TOP5}{\newline op\_scores}$
   \State $new\_graph \gets \Call{MUTATE}{initial\_graph, \newline final\_choose}$
    \State \textbf{return} $new\_graph$   
\EndProcedure

\end{algorithmic}
\label{alg_metamodel}
\end{algorithm}

\begin{figure*}
    \centering
    \includegraphics[width=0.8\linewidth]{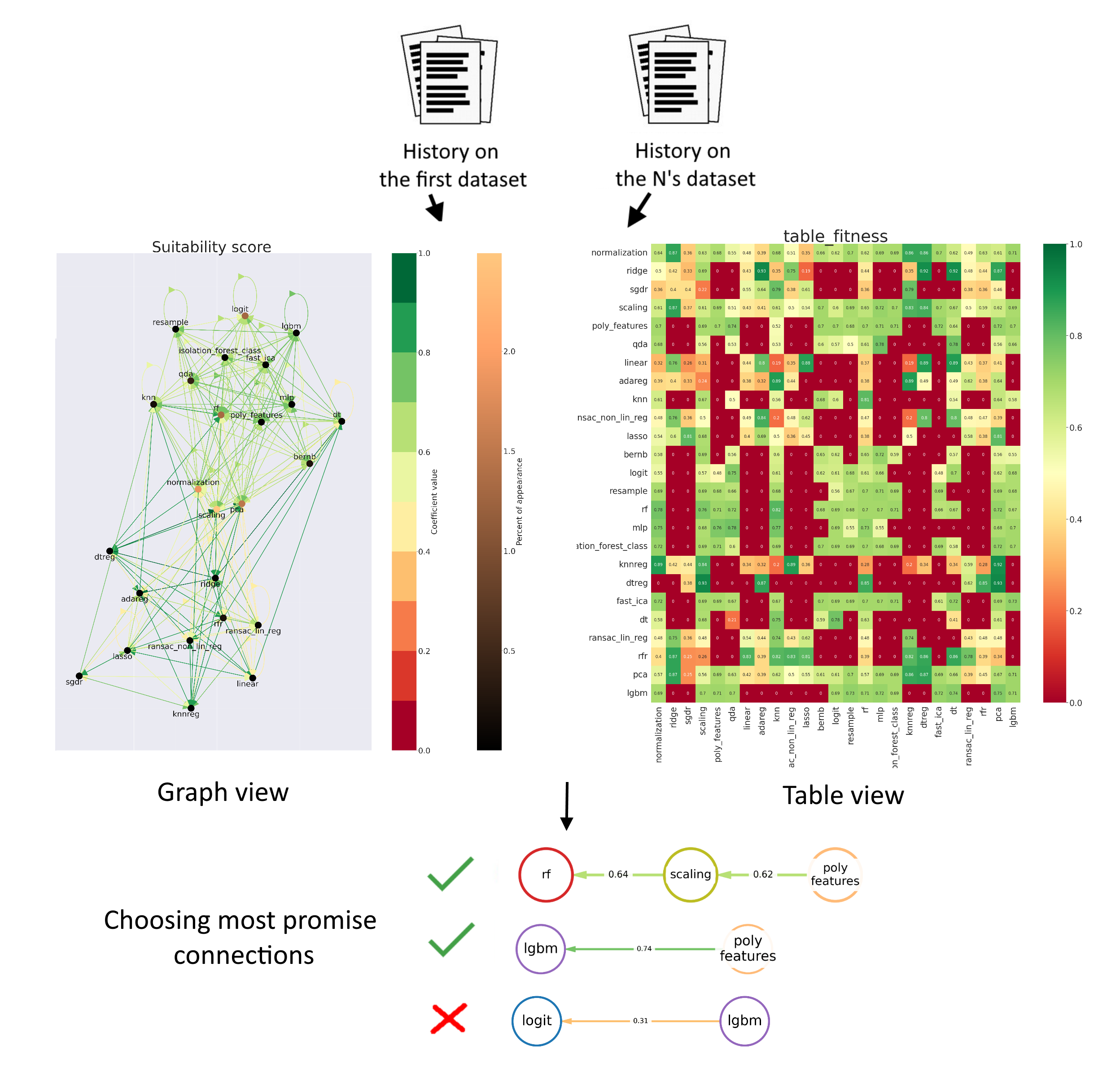}
    \captionsetup{justification=centering}
	\centering
    \caption{Workflow of the proposed approach to global SA (appearance analysis)}
    \label{fig_appearance}
\end{figure*}

The schema of the approach is presented in Figure~\ref{fig_appearance}.

\subsubsection{Criteria for analysis}
\label{subsec_target42}

The score between the operations $A$ and $B$ ($A\rightarrow B$) could be defined as 
$$ F_{suit}=\overline{Rn_{A\rightarrow B}}$$, where $\overline{Rn_{A\rightarrow B}}$ is the mean normalized rank of pipelines with the edge  $A \rightarrow B$.

The normalized rank $Rn$ is the pipeline place number in pipeline history, sorted by fitness divided by the number of pipelines in history.

\section{Experimental studies}
\label{sec_exp}

\subsection{Experimental setup}
\label{exp_setup}

The experiments are aimed to validate the hypothesis formulated in Section~\ref{sec_sa} and explored the efficiency and practical applicability of the EVOSA approach. These experiments focus on the local SA, since the full-scale evaluation of the meta-learning approach requires a more complicated set of experiments.

We use various datasets for binary and multi-class classification from AutoMLBenchmark \cite{gijsbers2019open}, and add other real-world datasets for multi-modal modelling and neural architecture search tasks.

F1, ROC AUC, and precision measures were used for the classification tasks; RMSE, MAE, MAPE, and $R^2$ were used for the regression tasks. The training time, inference time, and convergence time were analyzed as additional criteria for comparison.

The computational part of the experiments was conducted using Xeon Cascade Lake (2900MHz) with 12 cores and 24GB memory.

\subsection{Benchmarking}

To analyze the applicability of local SA techniques for individual cases, each dataset was analyzed separately. We compared  the F1 values for the proposed  \textit{EVOSA} approach and the alternative solutions - \textit{FEDOT} (without enhancements), \textit{AutoGluon} and \textit{H2O} (see Table~\ref{table:f1_mean}). We confirmed that \textit{FEDOT} selected as the foundation of the proposed approach is comparable with state-of-the-art solutions. Also, the SA component  improved the quality and reduced the number of failed runs (marked with dash). 

\begin{figure}[H]
    \centering
    \includegraphics[scale=0.3]{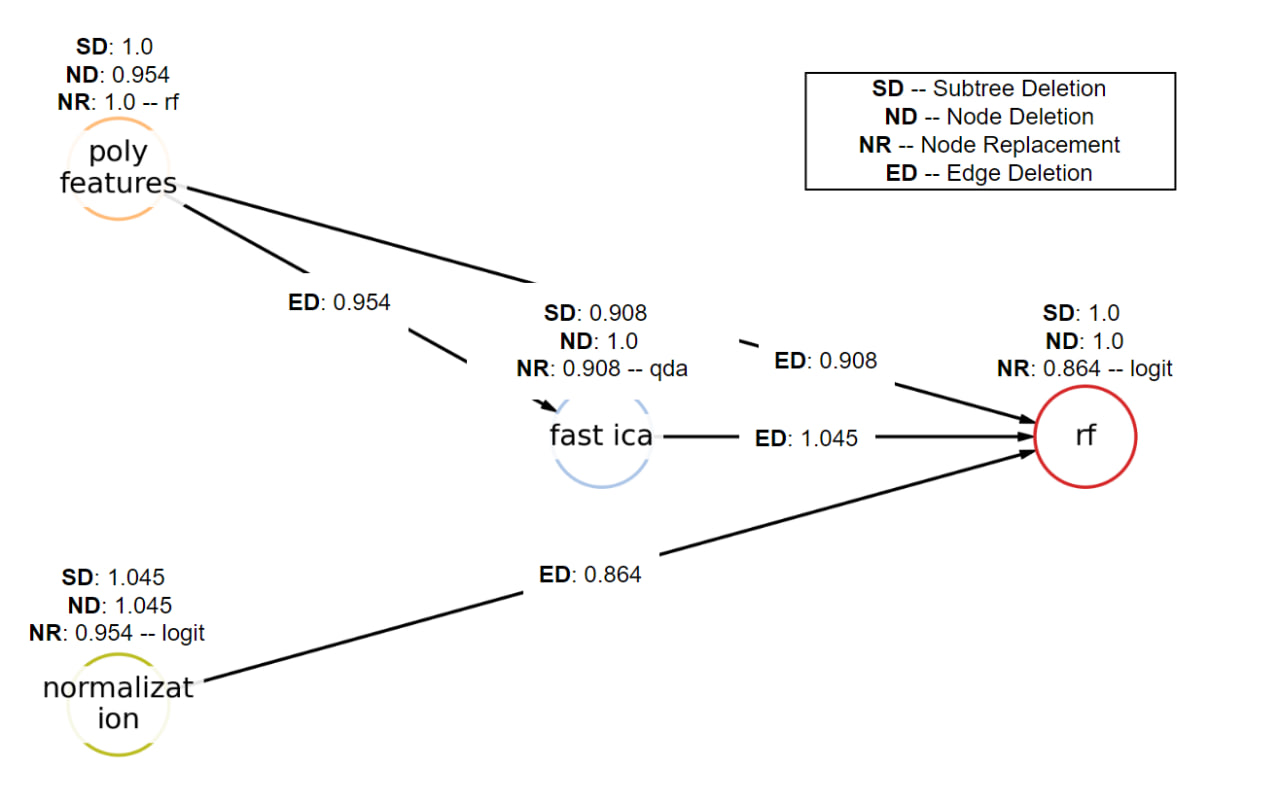}
	\centering
    \caption{Visualization made with sensitivity analysis for a graph on credit-g dataset.}
    \label{credit-g_SA}
\end{figure}

Figure~\ref{credit-g_SA} demonstrates an example of the SA output for a single dataset.

The original version of \textit{FEDOT} was compared to \textit{FEDOT } with \textit{EVOSA} in order to estimate the role of local SA in pipeline optimization. We observe that the SA component reduces the F1 variance by 6\% (see Table~\ref{table:f1_var}.). This indicates the increased stability of the AutoML with the SA component. When compared to \textit{AutoGluon} and \textit{H2O}, no significant differences were observed. This suggests that the SA brings greater benefits to solutions that prioritize variable-shaped pipelines.
The analysis of statistical significance for F1 metric in ~\ref{table:f1_var} confirms that  the local SA component does not compromise the predictive power of the pipelines.

The summary table in Table~\ref{table:inference_table} additionally confirms that the SA component successfully reduces the complexity of the final pipeline for the majority of the datasets.

%The statistical significance of the obtained results was also analyzed with the Wilcoxon test and it showed that the results are significant for 17 datasets. However, the results have improved and are also statistically significant for 5 datasets. 

The convergence time was also considered during the analysis. Reducing this time is extremely important, as it allows obtaining the results of the same quality more quickly. The results of the analysis are presented in Table~\ref{table:convergence_table}. Due to  local sensitivity analysis the average convergence time has decreased by 19.5\%, and was statistically significant for 6 datasets, where the training time decrease was consistently associated with the metric improvement. However, a significant increase in metrics and convergence was achieved only for small and medium datasets, advising caution about generalizing these patterns.

\begin{table*}[h!]
    \centering
    \caption{The summary table for all datasets with F1 metric. Failed runs are marked with dash.}
    \begin{tabular}{|c|c|c|c|c|}
    \hline
        \textbf{dataset} & \textbf{EVOSA} & \textbf{FEDOT} & \textbf{AutoGluon} & \textbf{H2O} \\ \hline
        \text{adult} & 0.874 & 0.874 & \textbf{0.875} & 0.874 \\ \hline
        \text{airlines} & 0.669 & 0.669 & \textbf{0.675} & 0.617 \\ \hline
        \text{airlinescodrnaadult} & 0.812 & - & \textbf{0.818} & 0.809 \\ \hline
        \text{albert} & 0.670 & 0.669 & \textbf{0.697} & 0.667 \\ \hline
        \text{amazon\_employee\_access} & 0.949 & 0.947 & 0.951 & \textbf{0.953} \\ \hline
        \text{apsfailure} & 0.994 & 0.994 & \textbf{0.995} & \textbf{0.995} \\ \hline
        \text{australian} & \textbf{0.871} & 0.870 & 0.865 & 0.860 \\ \hline
        \text{bank-marketing} & \textbf{0.910} & \textbf{0.910} & \textbf{0.910} & 0.899 \\ \hline
        \text{blood-transfusion} & 0.747 & 0.697 & \textbf{0.797} & 0.746 \\ \hline
        \text{car} & \textbf{1.000} & \textbf{1.000} & 0.998 & 0.998 \\ \hline
        \text{christine} & 0.746 & 0.746 & \textbf{0.748} & 0.737 \\ \hline
        \text{click\_prediction\_small} & \textbf{0.835} & \textbf{0.835} & 0.777 & 0.777 \\ \hline
        \text{cnae-9} & \textbf{0.957} & 0.954 & \textbf{0.957} & 0.954 \\ \hline
        \text{connect-4} & 0.792 & 0.788 & 0.865 & \textbf{0.867} \\ \hline
        \text{covertype} & 0.964 & 0.966 & \textbf{0.976} & 0.952 \\ \hline
        \text{credit-g} & 0.753 & 0.759 & \textbf{0.766} & 0.727 \\ \hline
        \text{dilbert} & 0.985 & 0.982 & \textbf{0.996} & 0.984 \\ \hline
        \text{fabert} & 0.688 & 0.685 & \textbf{0.939} & 0.534 \\ \hline
        \text{fashion-mnist} & \textbf{0.885} & - & 0.734 & 0.718 \\ \hline
        \text{guillermo} & 0.821 & - & \textbf{0.915} & 0.897 \\ \hline
        \text{helena} & 0.332 & 0.333 & \textbf{0.837} & 0.823 \\ \hline
        \text{higgs} & 0.731 & \textbf{0.732} & 0.369 & 0.336 \\ \hline
        \text{jannis} & 0.718 & 0.718 & \textbf{0.743} & 0.719 \\ \hline
        \text{jasmine} & 0.817 & \textbf{0.821} & 0.734 & 0.727 \\ \hline
        \text{jungle\_chess\_2pcs\_raw\_endgame\_complete} & \textbf{0.953} & 0.939 & 0.817 & 0.817 \\ \hline
        \text{kc1} & 0.866 & 0.867 & \textbf{0.996} & 0.947 \\ \hline
        \text{kddcup09\_appetency} & \textbf{0.982} & \textbf{0.982} & 0.866 & 0.818 \\ \hline
        \text{kr-vs-kp} & 0.995 & \textbf{0.996} & 0.982 & 0.962 \\ \hline
        \text{mfeat-factors} & \textbf{0.980} & 0.979 & \textbf{0.980} & \textbf{0.980} \\ \hline
        \text{miniboone} & 0.948 & 0.948 & \textbf{0.952} & 0.949 \\ \hline
        \text{nomao} & 0.969 & 0.970 & \textbf{0.975} & 0.974 \\ \hline
        \text{numerai28\_6} & \textbf{0.523} & 0.522 & 0.522 & 0.505 \\ \hline
        \text{phoneme} & 0.915 & \textbf{0.916} & \textbf{0.916} & 0.910 \\ \hline
        \text{riccardo} & 0.997 & - & \textbf{0.998} & 0.997 \\ \hline
        \text{robert} & 0.405 & - & \textbf{0.559} & 0.487 \\ \hline
        \text{segment} & \textbf{0.982} & \textbf{0.982} & \textbf{0.982} & 0.980 \\ \hline
        \text{shuttle} & \textbf{1.000} & \textbf{1.000} & \textbf{1.000} & \textbf{1.000} \\ \hline
        \text{sylvine} & \textbf{0.952} & 0.951 & \textbf{0.952} & 0.948 \\ \hline
        \text{vehicle} & \textbf{0.851} & 0.849 & 0.846 & 0.835 \\ \hline
        \text{volkert} & 0.694 & 0.694 & \textbf{0.758} & 0.697 \\ \hline
    \end{tabular}
    \label{table:f1_mean}
\end{table*}

\begin{table*}[!ht]
    \centering
    \caption{The summary table for all datasets for the standard deviation of F1 metric. The average improvement is near 6\%. 
    The statistical significance for F1 metric is also indicated.}
    \begin{tabular}{|c|c|c|c|c|}
    \hline
        \textbf{dataset} & \textbf{EVOSA} & \textbf{FEDOT} & \textbf{\% decline} & \textit{\textbf{p-value}} \\ \hline
        \textbf{amazon\_employee\_access} & 0.003 & 0.008 & 63.865 & 0.888 \\ \hline
        \textbf{connect-4} & 0.004 & 0.009 & 58.162 & 0.193\\ \hline
        \textbf{albert} & 0.003 & 0.005 & 44.172 & 0.211\\ \hline
        \textbf{sylvine} & 0.005 & 0.008 & 37.831 & 1.000\\ \hline
        \textbf{vehicle} & 0.023 & 0.032 & 25.919 & 1.000\\ \hline
        \textbf{higgs} & 0.005 & 0.007 & 20.406 & 0.906\\ \hline
        \textbf{blood-transfusion} & 0.085 & 0.107 & 20.305 & 0.593 \\ \hline
        \textbf{apsfailure} & 0.001 & 0.001 & 20.287 & 0.783\\ \hline
        \textbf{credit-g} & 0.026 & 0.030 & 13.887 & 0.598\\ \hline
        \textbf{kc1} & 0.015 & 0.017 & 12.939 & 0.753\\ \hline
        \textbf{numerai28\_6} & 0.004 & 0.004 & 11.302 & 0.820 \\ \hline
        \textbf{jungle\_chess\_2pcs\_raw\_endgame\_complete} & 0.038 & 0.042 & 9.793 & 0.767\\ \hline
        \textbf{dilbert} & 0.006 & 0.006 & 9.173 & 0.261\\ \hline
        \textbf{airlines} & 0.002 & 0.002 & 6.705 & 0.675\\ \hline
        \textbf{nomao} & 0.002 & 0.002 & 6.051 & 1.000\\ \hline
        \textbf{australian} & 0.029 & 0.030 & 5.903 & 0.833\\ \hline
        \textbf{covertype} & 0.005 & 0.005 & 5.182 & 0.678\\ \hline
        \textbf{bank-marketing} & 0.004 & 0.004 & 4.478 & 0.846\\ \hline
        \textbf{jasmine} & 0.020 & 0.020 & 2.444 & 0.695\\ \hline
        \textbf{shuttle} & 0.000 & 0.000 & 0.000 & 1.000\\ \hline
        \textbf{click\_prediction\_small} & 0.000 & 0.000 & 0.000 & 1.000\\ \hline
        \textbf{christine} & 0.018 & 0.018 & 0.000 & 0.799\\ \hline
        \textbf{jannis} & 0.004 & 0.004 & 0.000 & 1.000 \\ \hline
        \textbf{segment} & 0.008 & 0.008 & 0.000 & 1.000 \\ \hline
        \textbf{kddcup09\_appetency} & 0.000 & 0.000 & 0.000 & 1.000 \\ \hline
        \textbf{volkert} & 0.007 & 0.007 & 0.000 & 0.767 \\ \hline
        \textbf{miniboone} & 0.002 & 0.002 & -0.524 & 0.439\\ \hline
        \textbf{cnae-9} & 0.022 & 0.021 & -2.478 & 0.398\\ \hline
        \textbf{adult} & 0.004 & 0.004 & -6.305 & 0.922 \\ \hline
        \textbf{mfeat-factors} & 0.009 & 0.008 & -17.085 & 0.675 \\ \hline
        \textbf{kr-vs-kp} & 0.006 & 0.005 & -19.291 & 0.833\\ \hline
        \textbf{fabert} & 0.014 & 0.011 & -25.601 & 0.515\\ \hline
        \textbf{phoneme} & 0.019 & 0.015 & -27.117 & 1.000 \\ \hline
        \textbf{helena} & 0.023 & 0.013 & -71.389 & 1.000 \\ \hline
    \end{tabular}

    \label{table:f1_var}
\end{table*}

\begin{table*}[!ht]
    \centering
    \begin{tabular}{|c|c|c|c|c|}
    \hline
        \textbf{dataset} & \textbf{EVOSA} & \textbf{FEDOT} & \textbf{\% decline} & \textit{\textbf{p-value}} \\ \hline
        \textbf{credit-g} & \textbf{0.039} & \textbf{0.065} & \textbf{40.000} & \textbf{0.020} \\ \hline
        \text{blood-transfusion} & 0.009 & 0.012 & 25.000 & 0.625 \\ \hline
        \text{helena} & 0.577 & 0.708 & 18.503 & 0.695\\ \hline
        \textbf{shuttle} & \textbf{0.068} & \textbf{0.078} & \textbf{12.821} & \textbf{0.004} \\ \hline
        \text{sylvine} & 0.021 & 0.024 & 12.500 & 0.131\\ \hline
        \text{australian} & 0.019 & 0.021 & 9.524 & 0.652\\ \hline
        \text{jungle\_chess\_2pcs\_raw\_endgame\_complete} & 0.108 & 0.118 & 8.475 & 0.557 \\ \hline
        \textbf{amazon\_employee\_access} & \textbf{0.102} & \textbf{0.11} & \textbf{7.273} & \textbf{0.049}\\ \hline
        \textbf{kr-vs-kp} & \textbf{0.043} & \textbf{0.046} & \textbf{6.522} & \textbf{0.004}\\ \hline
        \textbf{cnae-9} & \textbf{0.416} & \textbf{0.443} & \textbf{6.095} & \textbf{0.016}\\ \hline
        \text{airlines} & 1.306 & 1.356 & 3.687 & 0.275\\ \hline
        \text{kc1} & 0.053 & 0.055 & 3.636 & 0.219\\ \hline
        \text{bank-marketing} & 0.237 & 0.242 & 2.066 & 0.084\\ \hline
        \text{nomao} & 1.257 & 1.26 & 0.238 & 0.557\\ \hline
        \text{phoneme} & 0.021 & 0.021 & 0.000 & 0.734 \\ \hline
        \textbf{segment} & \textbf{0.025} & \textbf{0.025} & \textbf{0.000} & \textbf{0.010}\\ \hline
        \text{connect-4} & 1.048 & 1.046 & -0.191 & 0.922\\ \hline
        \text{adult} & 0.255 & 0.254 & -0.394 & 0.770\\ \hline
        \text{jasmine} & 0.095 & 0.094 & -1.064 & 0.770\\ \hline
        \text{miniboone} & 0.749 & 0.736 & -1.766 & 0.275\\ \hline
        \textbf{higgs} & \textbf{0.765} & \textbf{0.749} & \textbf{-2.136} & \textbf{0.004}\\ \hline
        \textbf{numerai28\_6} & \textbf{0.217} & \textbf{0.212} & \textbf{-2.358} & \textbf{0.012}\\ \hline
        \textbf{fabert} & \textbf{1.87} & \textbf{1.825} & \textbf{-2.466} & \textbf{0.014}\\ \hline
        \textbf{jannis} & \textbf{0.52} & \textbf{0.507} & \textbf{-2.564} & \textbf{0.002}\\ \hline
        \textbf{christine} & \textbf{2.633} & \textbf{2.546} & \textbf{-3.417} & \textbf{0.006}\\ \hline
        \textbf{kddcup09\_appetency} & \textbf{1.877} & \textbf{1.790} & \textbf{-4.860} & \textbf{0.002} \\ \hline
        \text{click\_prediction\_small} & 15.418 & 14.551 & -5.958 & 0.084\\ \hline
        \textbf{mfeat-factors} & \textbf{0.098} & \textbf{0.092} & \textbf{-6.522} & \textbf{0.049}\\ \hline
        \text{dilbert} & 2.966 & 2.777 & -6.806 & 0.105\\ \hline
        \textbf{albert} & \textbf{14.231} & \textbf{12.018} & \textbf{-18.414} & \textbf{0.000}\\ \hline
        \text{car} & 0.016 & 0.013 & -23.077 & 0.084\\ \hline
        \textbf{covertype} & \textbf{7.582} & \textbf{5.449} & \textbf{-39.145} & \textbf{0.002}\\ \hline
        \text{vehicle} & 0.028 & 0.020 & -40.000 & 0.469\\ \hline
        \textbf{volkert} & \textbf{2.203} & \textbf{1.379} & \textbf{-59.753} & \textbf{0.002}\\ \hline
        \textbf{apsfailure} & \textbf{3.410} & \textbf{2.030} & \textbf{-67.980} & \textbf{0.002}\\ \hline
    \end{tabular}
    \caption{The summary table for all datasets for inference time metric. The results are sorted in order of inference duration reduction. The statistically significant results are highlighted.}
    \label{table:inference_table}
\end{table*}

\begin{table*}[!ht]
    \centering
    \begin{tabular}{|c|c|c|c|c|}
    \hline
        \textbf{dataset} & \textbf{EVOSA} & \textbf{FEDOT} & \textbf{\% decline} & \textit{\textbf{p-value}}\\ \hline
        \textbf{vehicle} & \textbf{1462.531} & \textbf{12647.102} & \textbf{88.436} & \textbf{0.031}\\ \hline
        \text{mfeat-factors} & 2766.071 & 18474.823 & 85.028 & 0.131\\ \hline
        \text{blood-transfusion} & 2346.217 & 12889.076 & 81.797 & 0.250\\ \hline
        \text{jasmine} & 5193.225 & 23060.815 & 77.480 & 0.922\\ \hline
        \textbf{credit-g} & \textbf{3740.598} & \textbf{13108.432} & \textbf{71.464} & \textbf{0.020}\\ \hline
        \textbf{cnae-9} & \textbf{4189.021} & \textbf{12772.144} & \textbf{67.202} & \textbf{0.016}\\ \hline
        \textbf{dilbert} & \textbf{2505.432} & \textbf{6156.280} & \textbf{59.303} & \textbf{0.049} \\ \hline
        \text{kc1} & 2600.527 & 6372.382 & 59.191 & 0.156\\ \hline
        \text{australian} & 2617.357 & 6214.310 & 57.882 & 0.496 \\ \hline
        \textbf{kr-vs-kp} & \textbf{10705.224} & \textbf{19728.782} & \textbf{45.738} & \textbf{0.027}\\ \hline
        \textbf{helena} & \textbf{13160.250} & \textbf{20251.298} & \textbf{35.015} & \textbf{0.002}\\ \hline
        \text{jungle\_chess\_2pcs\_raw\_endgame\_complete} & 8828.419 & 13543.298 & 34.813 & 0.232 \\ \hline
        \text{apsfailure} & 8013.792 & 11696.746 & 31.487 & 0.846 \\ \hline
        \text{phoneme} & 11797.471 & 16968.637 & 30.475 & 0.570\\ \hline
        \text{covertype} & 23090.240 & 30720.282 & 24.837 & 0.084\\ \hline
        \text{miniboone} & 10227.532 & 12929.883 & 20.900 & 0.432\\ \hline
        \text{bank-marketing} & 5903.615 & 7188.698 & 17.876 & 0.695 \\ \hline
        \text{segment} & 24170.218 & 28551.054 & 15.344 & 0.492 \\ \hline
        \text{shuttle} & 16798.069 & 19616.598 & 14.368 & 0.557 \\ \hline
        \text{connect-4} & 6922.035 & 7562.561 & 8.470 & 0.695\\ \hline
        \text{sylvine} & 16186.471 & 17548.978 & 7.764 & 0.492\\ \hline
        \text{airlines} & 14491.557 & 15569.290 & 6.922 & 1.000 \\ \hline
        \text{adult} & 7642.738 & 8094.004 & 5.575 & 0.322 \\ \hline
        \text{car} & 14671.193 & 15198.661 & 3.470 & 0.922 \\ \hline
        \text{christine} & 8631.138 & 8685.650 & 0.628 & 0.770 \\ \hline
        \text{nomao} & 5901.600 & 5930.465 & 0.487 & 0.770 \\ \hline
        \text{jannis} & 13009.647 & 12499.170 & -4.084 & 0.846 \\ \hline
        \text{amazon\_employee\_access} & 6042.829 & 5698.814 & -6.037 & 0.432 \\ \hline
        \text{volkert} & 11547.871 & 10650.005 & -8.431 & 1.000 \\ \hline
        \text{numerai28\_6} & 5553.752 & 4735.128 & -17.288 & 1.000 \\ \hline
        \text{albert} & 29824.311 & 23469.102 & -27.079 &  0.133 \\ \hline
        \text{higgs} & 8734.603 & 6834.383 & -27.804 & 0.625 \\ \hline
        \text{fabert} & 7197.197 & 5156.928 & -39.564 & 1.000 \\ \hline
        \text{kddcup09\_appetency} & 13099.101 & 5079.595 & -157.877 & 0.625 \\ \hline
    \end{tabular}
    \caption{The summary table for all datasets for convergence time metric. The results are sorted in descending order of decline. Average improvement is 19.5\%. The statistically significant results are highlighted.}
    \label{table:convergence_table}
\end{table*}

We observed a statistically significant improvement for the standard deviation of results, inference, and convergence time. The quality metrics are comparable with SOTA solutions. This confirms our hypothesis that the involvement of SA can improve the evolutionary AutoML without compromising the predictive power of obtained pipelines.

\begin{figure*}
    \centering
    \includegraphics[width=0.8\linewidth]{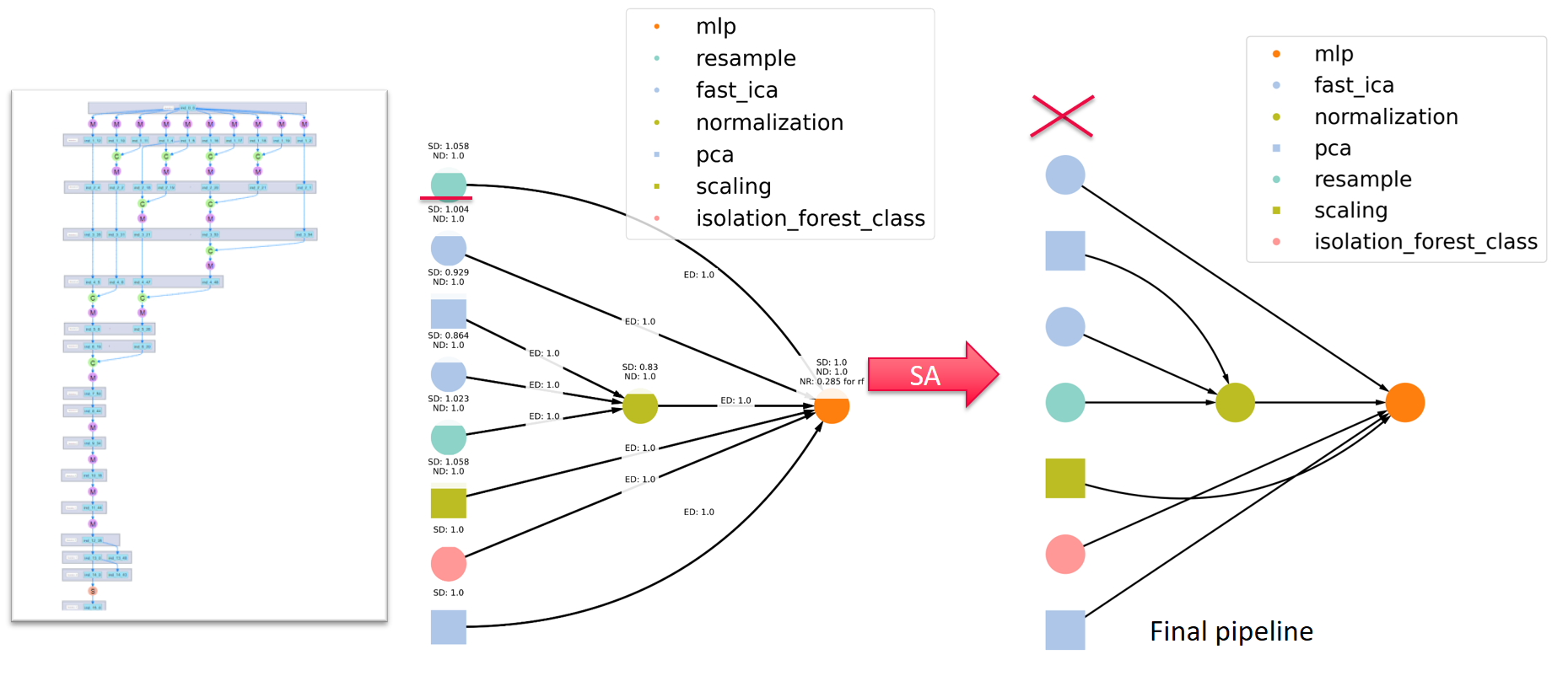}
    \captionsetup{justification=centering}
	\centering
    \caption{Visualization of evolutionary history for the jungle-chess dataset. Sensitivity Analysis is applied.}
    \label{fig_viz1}
\end{figure*}

In the implemented solution, each SA-based AutoML run can be visualized interactively. The visualization results are presented in Figure~\ref{fig_viz1}.

% \begin{figure*}
%     \centering
%     \includegraphics[width=0.8\linewidth]{images/ex_hist_2.png}
%     \captionsetup{justification=centering}
% 	\centering
%     \caption{Visualization of evolutionary history for the mfeat-factors dataset. Gluing-based SA mutation is applied.}
%     \label{fig_viz2}
% \end{figure*}

\begin{figure*}
    \centering
    \includegraphics[width=0.9\linewidth]{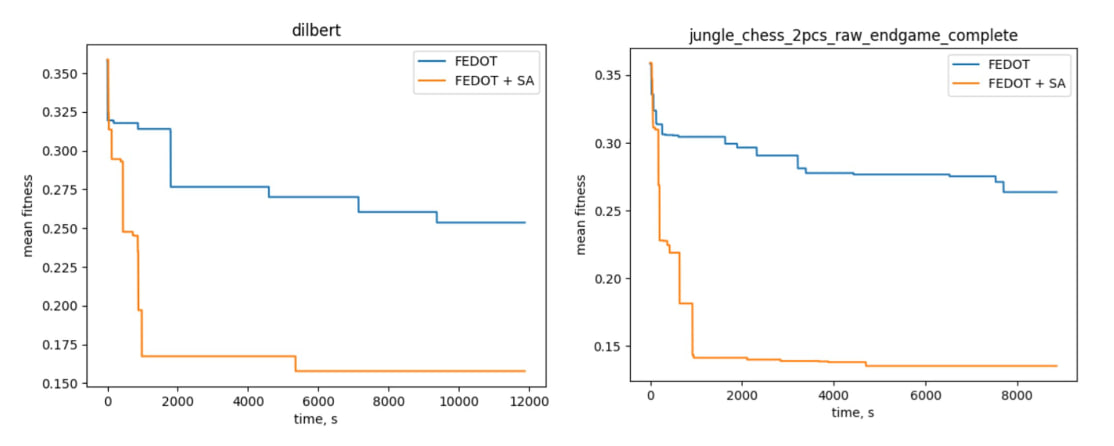}
    \captionsetup{justification=centering}
	\centering
    \caption{Visualization of the evolution process with the comparison of basic FEDOT and EVOSA. The convergence of fitness function is averaged for 10 runs.}
    \label{fig_fitness_per_time}
\end{figure*}

For a better understanding of how fitness improves in the process of evolution, the following visualizations are also presented in Figure~\ref{fig_fitness_per_time} Thus, it can be seen that the use of sensitivity analysis can significantly speed up the convergence process.

\begin{figure*}
    \centering
    \includegraphics[width=0.8\linewidth]{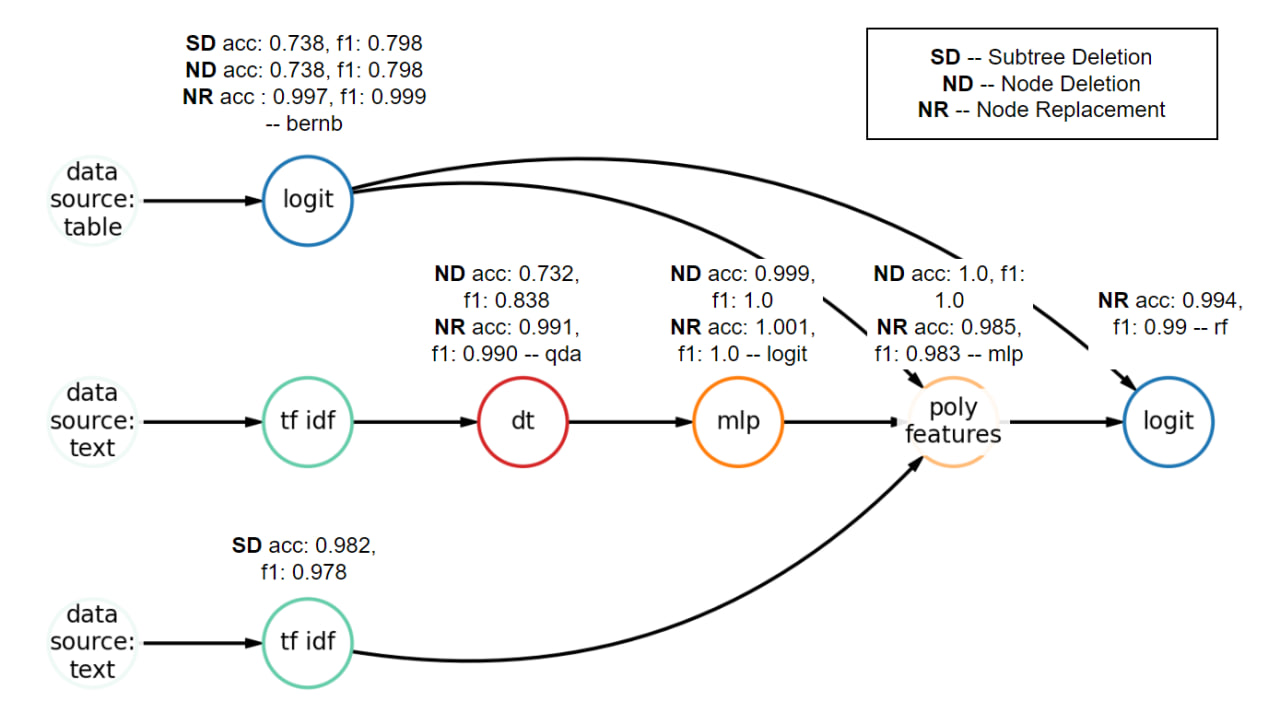}
    \captionsetup{justification=centering}
	\centering
    \caption{Example of visualization of multi-objective SA. The accuracy and f1 metrics are provided.}
    \label{fig_mo_sa}
\end{figure*}

Since the proposed sensitivity analysis techniques are implemented in a multi-objective way, several metrics can be used and visualized as a result. An example is provided in Figure~\ref{fig_mo_sa}, however, it has been simplified for greater readability.

\subsection{Real-world cases}

We explored the generalizability of the observed patterns, by testing the efficiency of EVOSA outside the well-known pull of AutoML tasks. Our extended experimental sample included a regression task for prediction of molecular energies, as well as multimodal classification and regression datasets. We additionally evaluated the local SA-based approach for the task of neural architecture search (NAS) for the convolutional neural networks (CNNs),

\subsubsection{SA for multimodal pipelines}

The experimental data were taken from the multimodal AutoML benchmark\footnote{\url{https://github.com/sxjscience/automl_multimodal_benchmark}}, which includes tabular datasets with textual columns. Due to limited computing power and large volumes of the benchmark, half of the datasets were used (the ratio of datasets' sizes and metrics was preserved). In some datasets, some columns were excluded (the same was done by benchmark authors)\footnote{\url{https://github.com/sxjscience/automl_multimodal_benchmark/blob/main/multimodal_text_benchmark/src/auto_mm_bench/datasets.py}}. Results of the application of local SA are presented in Table~\ref{table_mm_metric_comparison}. The obtained metrics indicate a significant advantage of EVOSA optimizer.

% The Example of analysis for \textit{prod} dataset is presented in Figure~\ref{fig_mm_sa}.

% \begin{figure*}
%     \centering
%     \includegraphics[width=0.8\linewidth]{images/mm_sa.png}
%     \captionsetup{justification=centering}
% 	\centering
%     \caption{Example of sensitivity analysis for the obtained pipeline}
%     \label{fig_mm_sa}
% \end{figure*}

\begin{table}[]
\begin{tabular}{|
>{\columncolor[HTML]{FFFFFF}}c |
>{\columncolor[HTML]{FFFFFF}}c |
>{\columncolor[HTML]{FFFFFF}}c |
>{\columncolor[HTML]{FFFFFF}}c |
>{\columncolor[HTML]{FFFFFF}}c |}
\hline
Dataset & Metric                     & \begin{tabular}[c]{@{}c@{}}Proposed \\ EVOSA \\ optimiser\end{tabular} & \begin{tabular}[c]{@{}c@{}}Auto\\ Gluon\end{tabular} & \begin{tabular}[c]{@{}c@{}}Impr., \\ \%\end{tabular} \\ \hline
jigsaw  & ROC-AUC                    & 0.947                                                                     & 0.967                                                & -2.07                                                \\ \hline
prod    & accuracy                   & 0.897                                                                     & 0.909                                                & -1.32                                                \\ \hline
qaa     & R2                         & 0.445                                                                     & 0.438                                                & +1.60                                                \\ \hline
qaq     & \cellcolor[HTML]{FFFFFF}R2 & 0.554                                                                     & 0.456                                                & +21.49                                               \\ \hline
mercari & \cellcolor[HTML]{FFFFFF}R2 & 0.999                                                                     & 0.605                                                & +65.12                                               \\ \hline
jc      & \cellcolor[HTML]{FFFFFF}R2 & 0.619                                                                     & 0.624                                                & -0.80                                                \\ \hline
average      & \cellcolor[HTML]{FFFFFF} &                                                                     &                                               & +14.00                                                \\ \hline
\end{tabular}
    \caption{Comparison of AutoGluon with EVOSA approach for the multimodal datasets.}
    \label{table_mm_metric_comparison}
\end{table}

\subsubsection{Molecular energy prediction}

The precise estimation of molecular energies is significant for reliable modelling of various chemical and biological processes in general and for solving different applied tasks; however, the application of existing analytical methods is highly computationally expensive and impractical \cite{laghuvarapu2020band}. Therefore, machine learning methods are becoming popular in this field due to their ability to learn general patterns in a given dataset and to predict properties of unknown structures, particularly the molecular energy\cite{unke2019physnet}. 

The fast development of deep neural networks allowed the researchers to make more accurate predictions of the energy of stable molecules. Most proposed models were evaluated on the QM9 dataset, which contains 134k small organic molecules in the potential energy stable (equilibrium) state\cite{ramakrishnan2014quantum}. In recent years, the best obtained MAE of these models was reduced from 0.0134 eV\cite{schutt2017schnet} to 0.0059 eV\cite{shui2020heterogeneous}.

The possibility of effective use of AutoML on molecular energy prediction was discovered in the MolHack 2022 competition organized by Artificial Intelligence Research Institute\footnote{\url{https://www.kaggle.com/competitions/molhack-2022/}}. Using EVOSA on raw data leads to the Top-3 results on the leaderboard. Models with better metrics are based on the sophisticated preprocessing and feature engineering, which require expert knowledge, and cannot be easily applied in an automated machine learning.

In this experiment, the performance of the SA-based evolutionary solution EVOSA was compared to three popular AutoML frameworks (\textit{AutoGluon}, \textit{H2O} and \textit{TPOT}). The training dataset contains 200,232 rows, split in a ratio of 0.75:0.25 (150,174 and 50,058 rows for train and test subsets), the validation dataset contained 32,884 rows. The frameworks were fitted under 1 hour and 4 hour restrictions, with the “best quality” preset (where possible), and on 12 concurrently running workers. No manual preprocessing was performed, and features did not contain information about molecular space structure. The results in Table~\ref{table:metric_comparison}
suggest that \textit{EVOSA} significantly outperforms \textit{AutoGluon} (AG) and \textit{H2O} frameworks with the 1-hour time limit; although RMSE of \textit{TPOT} is better in the 1-hour experiment, it does not improve with time (in the 4-hour experiment). In the 4-hour experiment, \textit{EVOSA} outperforms all competitors. All frameworks were fitted to raw data, and we expect additional profit for the AutoML system from the expert manual feature engineering and preprocessing.

\begin{table}[htbp]
    \centering
    \begin{tabularx}{\columnwidth}{|X|X|c|c|c|c|}
        \hline
            \textbf{Time}            & \textbf{Metric} & \textbf{AG} & \textbf{EVOSA} & \textbf{H2O}    & \textbf{TPOT}  \\ 
            \hline
            \multirow{2}{*}{1 h} & MAE    & 0.553     & \textbf{0.044} & 2.657  & 0.065 \\ \cline{2-6} 
                                 & RMSE   & 6.554     & 0.272          & 13.863 & \textbf{0.090} \\ \hline
            \multirow{2}{*}{4 h} & MAE    & 0.468     & \textbf{0.042} &        & 0.066 \\ \cline{2-6} 
                                 & RMSE   & 6.289     & \textbf{0.070} &        & 0.090 \\ \hline
    \end{tabularx}
    \caption{Comparison of AutoML solutions with EVOSA approach for the real-world regression dataset.}
    \label{table:metric_comparison}
\end{table}

% \begin{figure}
%     \centering
%     \includegraphics[width=0.8\linewidth]{images/molhack_pipeline.png}
%     \captionsetup{justification=centering}
% 	\centering
%     \caption{Example of sensitivity analysis for the obtained pipeline}
%     \label{fig_molhack_pipeline}
% \end{figure}

\textit{EVOSA} model inference is much faster than deep neural network models. However, \textit{EVOSA} cannot be used as a potential energy predictor due to a relatively high error. It can be used in a preliminary stage to classify molecules by their stability and filter non-acceptable ones to finally reduce the total time required for molecules' energy search\cite{schutt2017quantum}.

\subsubsection{NAS}

Finally, explored EVOSA outside the scope of ML pipeline design, applying it to neural architecture search for the CNNs. 

Our implementation of the NAS is based on the evolutionary optimizer described above. The structure of a CNN is represented as a graph. Each node corresponds to a neural network layer, and node parameters correspond to layer parameters such as the number of neurons and activation function for a dense layer, stride, padding, and pooling type for conv2d layers with different kernel sizes. Also, each node can have dropout and batch normalization layers as additional parameters. Thus, the proposed algorithm can generate deep CNNs using a variable number of different layers: conv2d 1x1, conv2d 3x3, conv2d 5x5 conv2d 7x7, batch normalization, dense, dropout, average and max pooling, and has support for skip connections with different shortcut lengths and the number of overlaps. 
% An example of the generated network is shown in Figure~\ref{fig_nas_graphs}.

% \begin{figure}[H]
%     \centering
%     \includegraphics[width=1.0\columnwidth]{images/new-nas-result.png}
%     \captionsetup{justification=centering}
% 	\centering
%     \caption{Example of NAS-generated graph for convolutions network}
%     \label{fig_nas_graphs}
% \end{figure}

The experiment was conducted with the butterfly classification dataset from Kaggle\footnote{\url{https://www.kaggle.com/datasets/gpiosenka/butterfly-images40-species}}, which contains about 10000 images of 75 different species of butterflies. The following parameters were used for the NAS experiments: population size - 15, number of generations - 15, number of epochs to train - 30, batch size - 16, and image size 256 with an ResNet-34 \cite{he2016deep} as initial assumption.

% The results comparison are described in Table~\ref{table:metric_comparison}.% and in  Figure~\ref{fig_nas_metrics}.

% \begin{table}[htbp]
%     \centering
%     \begin{tabularx}{\columnwidth}{|X|c|c|}
%         \hline
%          \textbf{Metric} & \textbf{NAS} & \textbf{Baseline}\\
%          \hline
%          LogLoss & 1.76 & 1.82\\ \hline
%          ROC-AUC & 0.982 & 0.97\\ \hline
%          Accuracy & 0.688 & 0.68\\ \hline
%     \end{tabularx}
%     \caption{Comparison of NAS-generated solution with baseline}
%     \label{table:metric_comparison}
% \end{table}

\begin{figure}[H]
    \centering
    \includegraphics[width=0.95\linewidth]{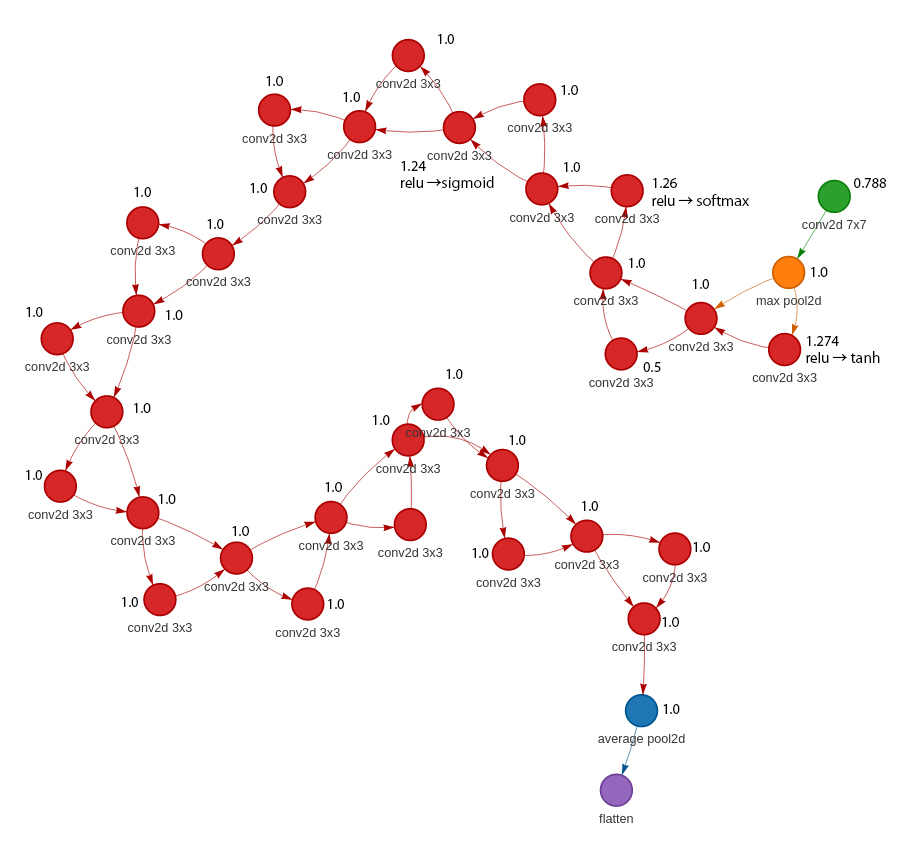}
    \captionsetup{justification=centering}
	\centering
    \caption{NAS-generated graph after local SA.}
    \label{nas-sa-graph}
\end{figure}

%\subsubsection{Sensitivity analysis for NAS}
Due to the neural network's graph representation, we can also use sensitivity analysis for further adjustments to the architecture of CNN. We can analyze each node and each edge of the neural network in terms of how the fitness function changes if we apply transformations to the nodes or edges. It allows optimizing the neural network architecture, dropping or replacing redundant layers or connections.

The experiment took about 8 hours. After applying sensitivity analysis, we increased the neural network's performance on the same task by changing the parameters of several layers. This resulted in an optimized architecture similar to NAS in terms of performance, but with an improved convergence and inference durations (see Table~\ref{table:nas-sa-res} and Figure~\ref{nas-sa-graph})

\begin{table}[H]
    \centering
    \begin{tabularx}{\columnwidth}{|X|c|c|}
        \hline
         \textbf{Metric} & \textbf{Before SA} & \textbf{After SA}\\
         \hline
         Fitness & 5.53 & 3.82\\ \hline
         ROC-AUC & 0.974 & 0.987\\ \hline
         Accuracy & 0.586 & 0.713\\ \hline
         logloss & 2.156 & 1.276\\ \hline
    \end{tabularx}
    \caption{Application of SA to the results of NAS}
    \label{table:nas-sa-res}
\end{table}

% \begin{figure}[H]
%     \centering
%     \includegraphics[width=0.8\linewidth]{images/sa-result.jpg}
%     \captionsetup{justification=centering}
% 	\centering
%     \caption{NAS-generated graph after local SA.}
%     \label{nas-sa-graph}
% \end{figure}

% \begin{figure}[h]
%     \begin{subfigure}{0.45\textwidth}
%         \includegraphics[width=1\linewidth]{images/NAS_acc.png}
%         \caption{NAS accuracy}
%         \label{fig_nas_acc}
%     \end{subfigure}
%     \begin{subfigure}{0.45\textwidth}
%         \includegraphics[width=1\linewidth]{images/resnet_acc.png}
%         \caption{ResNet accuracy}
%         \label{fig_res_acc}
%     \end{subfigure} \\[\smallskipamount]
%     \begin{subfigure}{0.45\textwidth}
%         \includegraphics[width=1\linewidth]{images/NAS_loss.png}
%         \caption{NAS loss}
%         \label{fig_nas_loss}
%     \end{subfigure}
%     \begin{subfigure}{0.45\textwidth}
%         \includegraphics[width=1\linewidth]{images/resnet_loss.png}
%         \caption{ResNet loss}
%         \label{fig_res_loss}
%     \end{subfigure} \\[\smallskipamount]
%     \caption{Metric by epoch}
%     \label{fig_nas_metrics}
    
% \end{figure}

% results

\section{Discussion}
\label{sec_disc}

In this study, we aimed to confirm that the integration of SA and evolutionary optimization can increase the effectiveness of AutoML. The experimental study was designed to analyze this hypothesis from different sides. The set of benchmarks and real-world cases were involved in comparison, as well as various criteria were considered: quality-based, stability-based, convergence-based, and time-based.

The obtained results were analyzed to confirm or reject the hypothesis formulated in Section~\ref{sec_ps}. First of all, there is no statistically significant difference in prediction quality metrics between the EVOSA and state-of-the-art solutions (FEDOT, AutoGluon, H2O). However, the predictive quality highly depends on the set of “building blocks” (models and operations) used during optimization. Also, the relatively small number of datasets in AutoMLBenchmark and values of quality measures that are already close to optima make the use of statistical tests not that illustrative.

For this reason, we are focusing the analysis not on the comparison between state-of-the-art solutions itself, but on the analysis of improvement in convergence time and decreased variance of results and complexity of pipelines after the involvement of SA in evolutionary optimization. This part of the analysis confirms the practical applicability of SA techniques, since the faster convergence is quite critical for AutoML tool \cite{lazebnik2022substrat}. Also, we successfully investigated the applicability of EVOSA to real-world cases and tasks from different fields: from multi-modal modelling to neural architecture search. So, we can conclude the hypothesis in empirically confirmed (since convergence time, inference time and standard deviation of quality metrics and reduced without loss in predictive quality).

We used FEDOT AutoML framework as a case study in the paper. However, the EVOSA can be applied for other AutoML tools that face the similiar problems (over-complicated and non-explainable pipelines).

The future extension of this research can be directed to the full-scale implementation of meta-learning techniques based on global SA and its extension to other AutoML tasks (e.g. time series forecasting and time series classification). It requires both implementation of this approach and an extensive set of experiments to confirm its applicability. Also, the surrogate models based on graph neural networks can be used to reduce the computational cost of SA.

\section{Conclusion}
\label{sec_concl}

This paper implements the techniques for integrating evolutionary AutoML and SA. We propose the EVOSA approach that can be used to design variable-shaped pipelines. Besides their theoretical applicability, the setup for experimental validation is described. 

The following datasets are used for experiments:
(1) Tabular datasets from OpenML (classification/regression); 
(2) Tabular from hackathon (regression); 
(3) Multimodal datasets from the benchmark (classification/regression); 
(4) Image dataset from Kaggle (classification).

The achieved results can be explained in the following way. 
For the AutoMLBenchmark, the quality metrics do not significantly differ from the state-of-the-art solutions. However, the improvement in convergence time is near 20\%, as well as a 5\% improvement in the stability of results. Also, statistical significance was achieved for 15 datasets by inference time and for 6 datasets by convergence time. 
It was also noted that local SA tends to work better on small and medium datasets. It is important to note that EVOSA avoids running structural optimization if the training time for the model is comparable with a full time limit.

Also, for more complicated real-world cases, the improvement exceeds 5\%. Also, we run the experiments to confirm the applicability of implemented solution not only for ML pipelines, but also for search of optimal architectures of convolutional neural networks. For these reasons, EVOSA can be considered a valuable tool set for the interpretation of the data-driven models (both ML pipelines and neural networks). As an extension of this research, it is promising to conduct the wide-scale set of experiments for global SA to evaluate the applicability of SA-based meta-learning. The experimental studies should involve not only tabular classification and regression tasks, but also time series forecasting and time series classification as more complicated and diverse problems.

% \section{Acknowledgements}

% This work was supported by the Analytical Center for the Government of the Russian Federation (IGK 000000D730321P5Q0002), agreement No. 70-2021-00141.

\section{REFERENCES}
\printbibliography[heading=none]

\end{document}